\documentclass[letterpaper, 10 pt, journal, twoside]{ieeetran}

\usepackage[noadjust]{cite}

\usepackage[utf8]{inputenc} 
\usepackage[T1]{fontenc}    
\usepackage{hyperref}       
\usepackage{url}            
\usepackage{booktabs}       
\usepackage{amsfonts}       
\usepackage{nicefrac}       
\usepackage{microtype}      
\usepackage{xcolor}         
\usepackage{graphicx}
\usepackage{amssymb}
\usepackage{amsmath}
\usepackage[detect-weight]{siunitx}
\usepackage{multirow}
\usepackage{algorithm}
\usepackage{algorithmic}
\usepackage{subcaption}
\usepackage{hhline}
\usepackage{mathtools}

\usepackage{enumitem}
\usepackage{siunitx}
\usepackage[usestackEOL]{stackengine}
\usepackage{multirow}
\usepackage{soul}
\usepackage{ulem}

\usepackage{pgfplots}
\DeclareUnicodeCharacter{2212}{−}
\usepgfplotslibrary{groupplots,dateplot}
\usetikzlibrary{patterns,shapes.arrows}
\pgfplotsset{compat=newest}

\title{Athletic Mobile Manipulator System for Robotic Wheelchair Tennis}

\author{Zulfiqar Zaidi$^{*}$, Daniel Martin$^{*}$, Nathaniel Belles, Viacheslav Zakharov, Arjun Krishna, Kin Man Lee,\\ Peter Wagstaff, Sumedh Naik, Matthew Sklar, Sugju Choi, Yoshiki Kakehi, Ruturaj Patil,\\Divya Mallemadugula, Florian Pesce, Peter Wilson, Wendell Hom, Matan Diamond, Bryan Zhao,\\ Nina Moorman, Rohan Paleja, Letian Chen, Esmaeil Seraj, and Matthew Gombolay,~\IEEEmembership{Member,~IEEE}
\thanks{Manuscript received: October, 2, 2022; Revised: December, 20, 2022; Accepted: January, 15, 2023.}
\thanks{This letter was recommended for publication by Editor Tamim Asfour upon evaluation of the reviewers' comments.
This work was supported by a gift from Google and Georgia Institute of Technology's Ken Byers Tennis Complex.} 
\thanks{$^{*}$Equal Contribution}
\thanks{The authors are with the Georgia Institute of Technology, Atlanta, GA 30332, USA. Corresponding email:      {\tt\footnotesize zzaidi8@gatech.edu}}%
\thanks{Digital Object Identifier (DOI): see top of this page.}
}

\begin{document}

\maketitle

\begin{abstract}

Athletics are a quintessential and universal expression of humanity. From French monks who in the $12^{th}$ century invented \emph{jeu de paume}, the precursor to modern lawn tennis, back to the \emph{K'iche'} people who played the Maya Ballgame as a form of religious expression over three thousand years ago, humans have sought to train their minds and bodies to excel in sporting contests. Advances in robotics are opening up the possibility of robots in sports. Yet, key challenges remain, as most prior works in robotics for sports are limited to pristine sensing environments, do not require significant force generation, or are on miniaturized scales unsuited for joint human-robot play. In this paper, we propose the first open-source, autonomous robot for playing regulation wheelchair tennis. We demonstrate the performance of our full-stack system in executing \emph{ground strokes} and evaluate each of the system's hardware and software components. The goal of this paper is to (1) inspire more research in human-scale robot athletics and (2) establish the first baseline for a reproducible wheelchair tennis robot for regulation singles play. Our paper contributes to the science of systems design and poses a set of key challenges for the robotics community to address in striving towards robots that can match human capabilities in sports.
\end{abstract}


\section{Introduction}
\label{sec:intro}
\IEEEPARstart{S}{ports} have been an integral part of human history as they have served as an important venue for humans to push their athletic and mental abilities. Sports transcend culture~\cite{taylor_2013,Seippel_cross_national_sports_importance} and provide strong physiological and social benefits~\cite{malm2019physical}. The physical and mental training from sports is widely applicable to multiple aspects of life~\cite{Eime2013,Malm2019-fv}. Developing robotic systems for competitive sports can increase participation in sports and be used for sports training~\cite{Reinkensmeyer2009-ex,biomechanics_sports}, thus benefiting society by promoting a healthier lifestyle and increased economic activity. Furthermore, sports serve as a promising domain for developing new robotic systems through the exploration of high-speed athletic behaviors and human-robot collaboration.

Researchers in robotics have sought to develop autonomous systems for playing various sports such as soccer and table tennis. RoboGames~\cite{calkins2011overview} and RoboCup~\cite{carbonell_robocup_1998} have inspired the next generation of roboticists to solve challenges in sensing, navigation, and control. However, many of the advancements or techniques used in solving these obstacles have been restricted to overly miniaturized (e.g. Robo-Soccer~\cite{cass2001robosoccer}, Robo-sumo~\cite{robot_sumo}), relatively stationary~\cite{muelling2010learning}, or highly idealized/controlled environments~\cite{hattori-humanoidtennisswing2020,terasawa-humanoidtennisswing2016,robot_sumo}.
The hardware is often specifically engineered to the sport~\cite{buchler2022learning}, making re-purposing designs and components for other applications difficult. We address these limitations of prior work by proposing an autonomous system for regulation wheelchair tennis by leveraging only commercial off-the-shelf (COTS) hardware. Our system is named ESTHER (\underline{E}xperimental \underline{S}port \underline{T}ennis w\underline{HE}elchair \underline{R}obot) after arguably the greatest wheelchair tennis player, Esther Vergeer.

\begin{figure}
    \centering
    \includegraphics[width=0.975\linewidth, height = 7 cm]{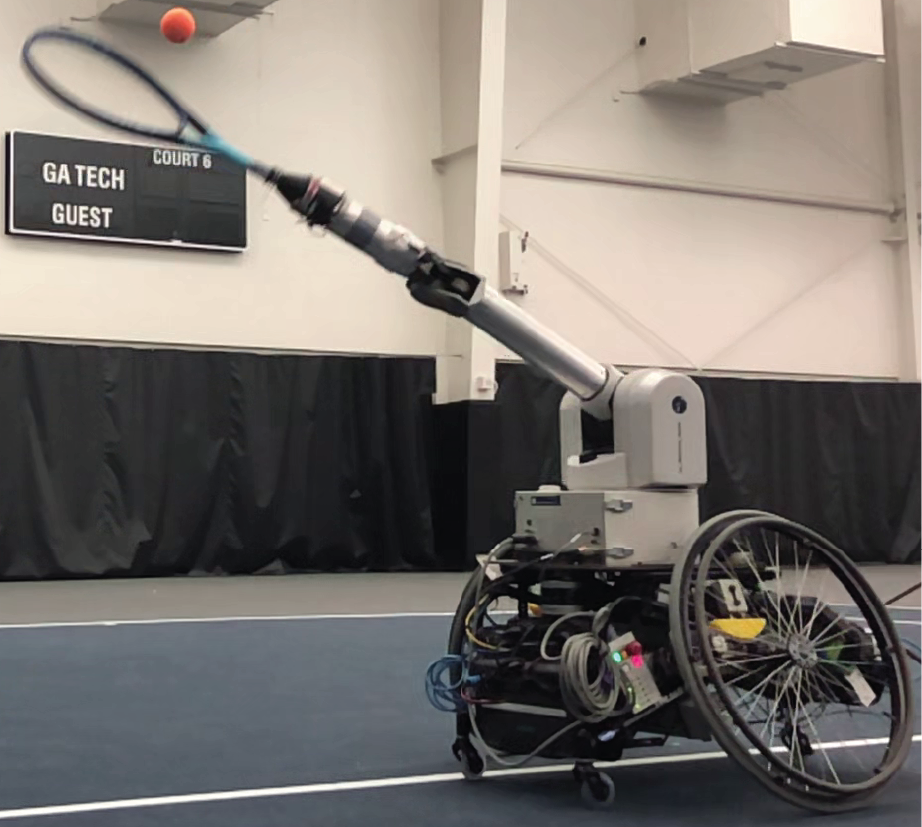}
    \vspace{-4pt}
    \caption {ESTHER hitting a tennis ground stroke.}
    \label{fig:wheelchair-robot}
        \vspace{-8pt}
\end{figure}

Tennis is a challenging sport requiring players to exhibit accurate trajectory estimation, strategic positioning, tactical shot selection, and dynamic racket swings. The design of a system that meets these demands comes with a multitude of complications to address: precise perception, fast planning, low-drift control, and highly-responsive actuation. These challenges must be resolved in a framework efficient enough to respond in fractions of a second~\cite{tennis_timepressure}. 

In this paper, we (1) present the system design that enables ESTHER (Fig. 1) to address these challenges, and (2) conduct an empirical analysis to evaluate the capabilities of our system. Our design adheres to ITF tennis regulations by utilizing a legal tennis wheelchair and an anthropomorphic robot arm, which were required to obtain permission to field ESTHER on an NCAA Division 1 indoor tennis facility. We contribute to the science of robotic systems by (1) Demonstrating a fully-mobile setup including a decentralized low-latency vision system and a 86 kg mobile manipulator that can be set up within 30 minutes, (2) Presenting a design for a motorized wheelchair that meets the athletic demands of tennis under the full-load of the manipulator, onboard computer, and batteries, and (3) Using our mobile manipulator system, and a decentralized low-latency vision system made using 6 low-cost passive 3D cameras to establish the following baselines for robotic wheelchair tennis: asynchronous ball detection up to a maximum rate of 150 Hz; successful return rate of up to 66$\%$ for balls launched at a slow speed in an indoor tennis court; achieve a racket-head speed of \SI{10}{\meter\per\second} while hitting a ground stroke; achieve speeds and accelerations of up to \SI{5}{\meter\per\second} and \SI{1}{\meter\second\tothe{-2}}, respectively, with a wheelchair base.

We open-sourced our designs and software on our project website: \url{https://core-robotics-lab.github.io/Wheelchair-Tennis-Robot/}

\begin{figure}
     \centering
        \includegraphics[width=0.82\linewidth]{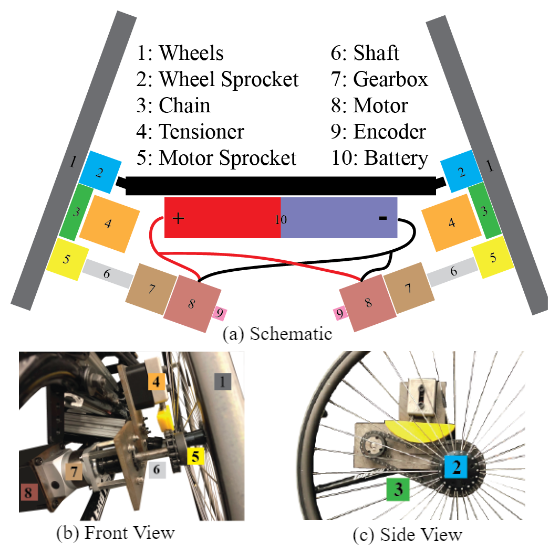}
     \vspace{-7.5pt}
        \caption{The schematic (Fig. a) and images (Fig. b-c) of the mechanical assembly of the chain drive system.}
        \label{fig:wheelchair-mechanical}
\end{figure}

\section{Related Work}

Systems that play tennis must be able to rapidly traverse the court, control the racket at high speeds with precision, and deliver sufficient force to withstand the impulse from ball contact. Existing systems that attempt to play other racket sports~\cite{mori-badminton-pneumatic2019,buchler2022learning} typically use pneumatically driven manipulators for quick and powerful control; however, these systems are stationary and are thus unsuitable.
Other works~\cite{janssens-badmintonrail2012,liu-badmintonlearning2013,gao-deeprlrailtabletennis2020,zhang-humanoidrail2014,miyazaki-railtabletennis2006,ji-railtabletennis2021} have mounted robotic manipulators on rails, but these systems do not support full maneuverability. Furthermore, these robot environments are heavily modified to suit the robot, rather than enabling the robot to act more naturally within the existing human environment.

There have been several recent works in creating agile mobile manipulators~\cite{dong_catch_2020,yang-varsm2022}. For example,~\cite{dong_catch_2020} proposes a UR10 arm with 6 degrees of freedom (DoF) mounted on a 3-DoF omni-directional base for catching balls. This system works well for low-power, precision tasks, but lacks the force required for a tennis swing. VaRSM~\cite{yang-varsm2022} is a system designed to play racket sports including tennis and is most similar to our work. The system proposed uses a custom 6-DoF arm and custom swerve-drive platform as its base. While this system appears effective, it is difficult to reproduce given that the hardware is custom-made. Further, the system was developed by a company that has not open-sourced its implementation, and experimental details are lacking to afford a proper baseline. In contrast, ESTHER is a human-scale robot constructed from a COTS 7-DoF robot manipulator arm and a regulation wheelchair with an open-sourced design and code-base. This enables our system to serve as a baseline for human-scale athletic robots while being reproducible.

\begin{figure}
\includegraphics[width=\linewidth]{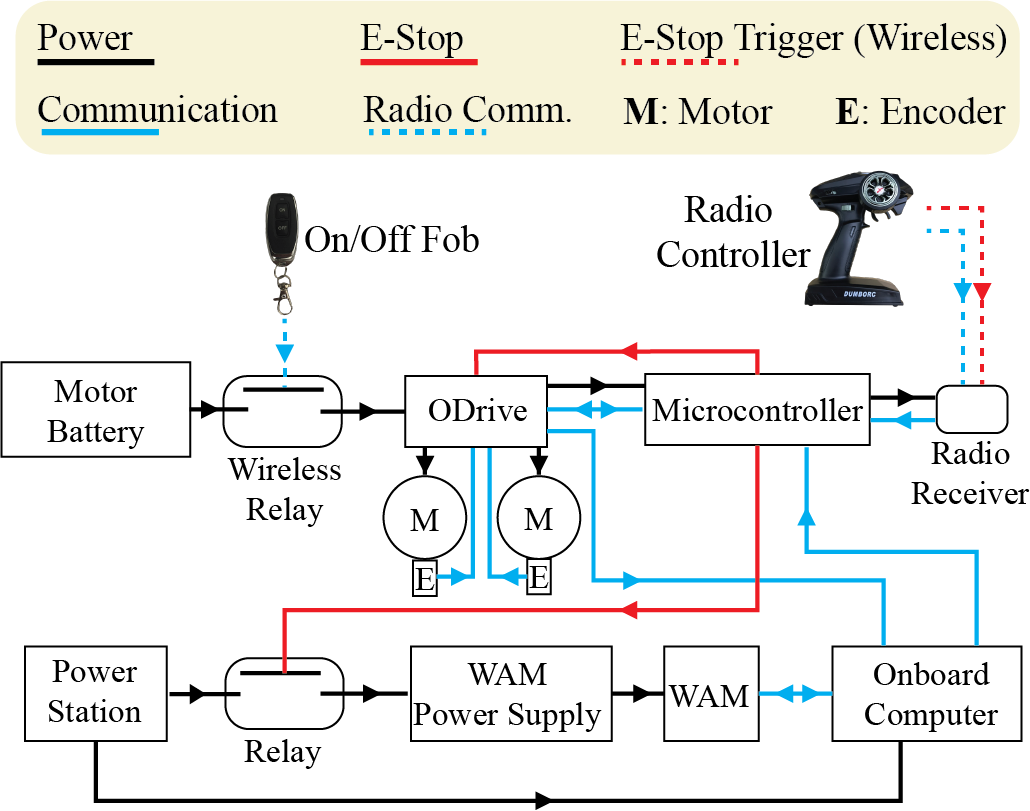}
    \caption {Electrical and communication diagram for ESTHER's mobile base and robotic arm.}
    \label{fig:wheelchair-electrical}
\end{figure}

\section{METHODS}
\label{sec:methods}

We detail the ESTHER's hardware (Sec.~\ref{subsec:hardware}), perception (Sec.~\ref{subsubsec:sensing}), planning (Sec.~\ref{subsubsec:planning}), and control (Sec.~\ref{subsubsec:control}) components. ESTHER's design meets the athletic demands of tennis and is easy and quick to set up. 

\vspace{-3mm}
\subsection{Hardware}
\label{subsec:hardware}

We mounted a 7-DoF high-speed Barrett WAM on a motorized Top End Pro Tennis Wheelchair (Fig.~\ref{fig:wheelchair-robot}).

\textbf{Wheelchair} --
We designed a chain-drive system to deliver power from the motors to the wheels. The schematic of the system's mechanical assembly is illustrated in Fig.~\ref{fig:wheelchair-mechanical}. Two motors for both wheels are coupled to a 1:10 ratio speed-reducer planetary gearbox. The gearbox output shaft is coupled with the wheel through a chain and sprocket system that provides an additional 1:2 speed reduction to give a total reduction of 1:20.
With the motors rotating at maximum speed, the wheelchair can achieve linear velocities of up to \SI{10}{\meter\per\second} and in-place angular yaw velocity of up to \SI{20}{\radian\per\second}. A chain drive system has higher durability, torque capacity, increased tolerances, and a simpler design versus belt drive or friction drive systems.

\begin{figure}
    \includegraphics[width=\linewidth]{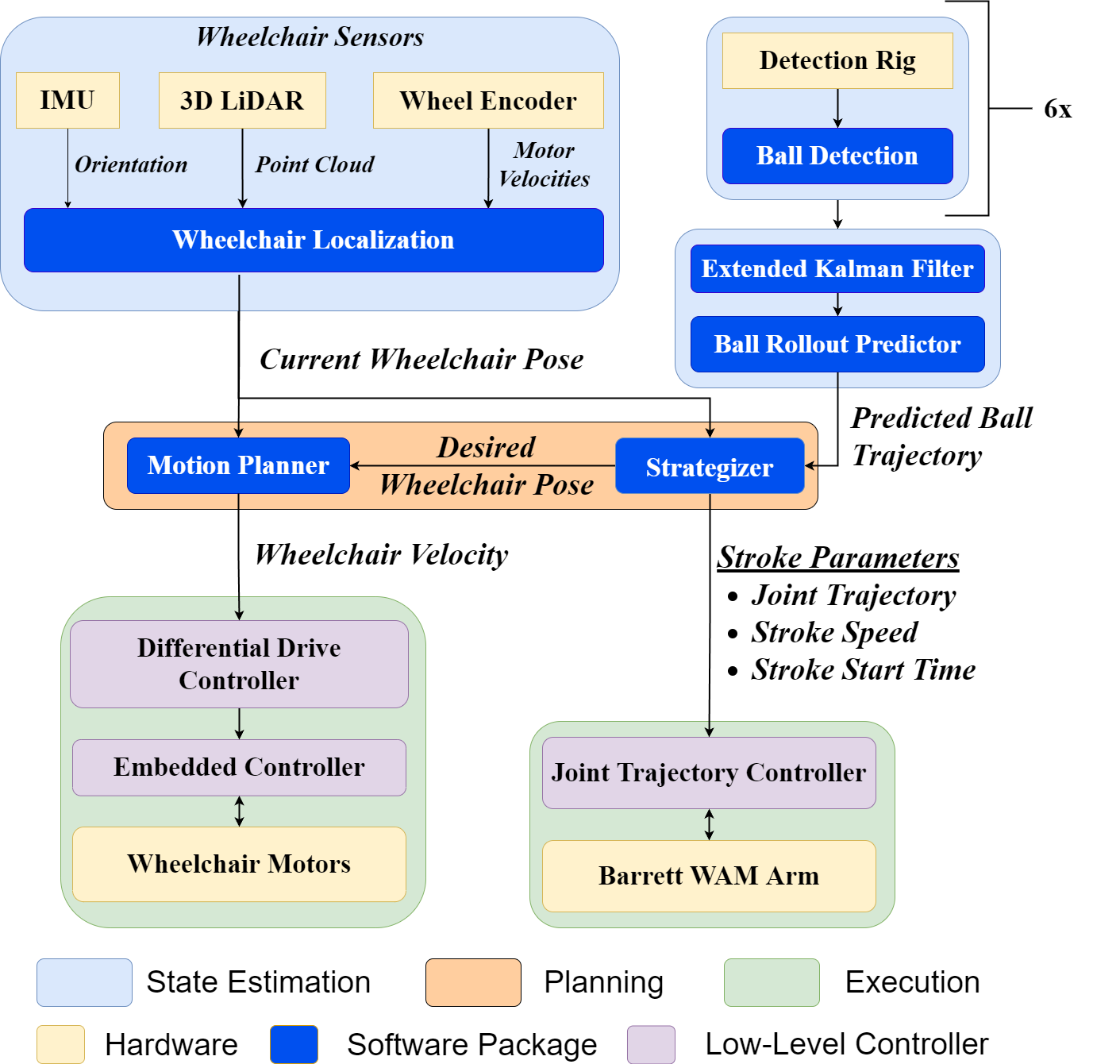}
    \caption{ESTHER's System Architecture.}
    \label{fig:architecture}
\end{figure}
The electrical system to power and control the mobile wheelchair base consists of a DC battery, two ODrive D6374 motors, an ODrive motor controller, a wireless relay, a wireless radio controller and receiver pair, and a microcontroller as illustrated in Fig.~\ref{fig:wheelchair-electrical}. The wireless relay allows the user to remove all power from the drive system and WAM, and acts as the main safety mechanism to stop all actuators. The radio controller allows the user to remotely swap between idle, manual, and autonomous modes. In idle mode, motors are de-energized. In manual mode, a human can remotely control the motion of the wheelchair. In autonomous mode, the wheelchair's motion is controlled by the onboard computer that communicates the desired wheel velocities to the microcontroller. The radio controller also has an emergency stop (E-stop) button, which acts as a second safety mechanism to remove power from the WAM and command the drive motors to stop moving. A third safety mechanism is implemented in software on the microcontroller that stops all motion if it stops receiving new commands from the computer. Maintaining these safety mechanisms enables the whole system to be tether-free and safely operable from a distance. As for electrical dangers, all bare wire terminations are enclosed in a small box to ensure the safety of any operator while powered on.

\textbf{Barrett WAM} --
We mount a ``HEAD Graphene Instinct Power'' tennis racket at the end of the WAM arm using a 3D-printed connector (Fig.~\ref{fig:wheelchair-robot}). The design and visualization of potential failure modes are available on the project website. The arm is powered by a Anker 535 \SI{500}{\watt} portable power station that has standard \SI{110}{\watt} AC outlets as illustrated in Fig.~\ref{fig:wheelchair-electrical}. The WAM power supply pulls power from one of the AC outlets and supplies \SI{80}{\volt} DC power over a cable terminated with locking connectors to the WAM. This specific voltage is required to get the maximum joint speeds needed for this application. The power station is also used to power the onboard computer, the LIDAR, and the WAM making the whole system tether-free. The computer requires up to \SI{300}{\watt} when running the full code stack. The LIDAR requires \SI{30}{\watt} when running at maximum frequency. The WAM requires \SI{50}{\watt} when static and up to \SI{150}{\watt} while swinging. 

\vspace{-4mm}
\subsection{System Design}
\label{subsec:architecture}

We decompose the ESTHER into three main components: Sensing and State Estimation, Planning, and Low-Level Controllers (Fig.~\ref{fig:architecture}). We leverage ROS \cite{ros, ros-guide} to build an interconnected modular software stack using a combination of open-source packages \cite{ros-guide} and custom packages.

\subsubsection{Sensing and State Estimation}
\label{subsubsec:sensing}

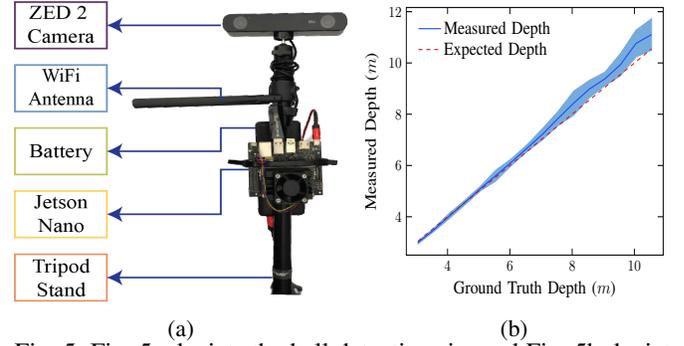
\begin{figure}
     \centering
     \begin{subfigure}[b]{0.5\linewidth}
    \includegraphics[width=\linewidth, height = 4cm]{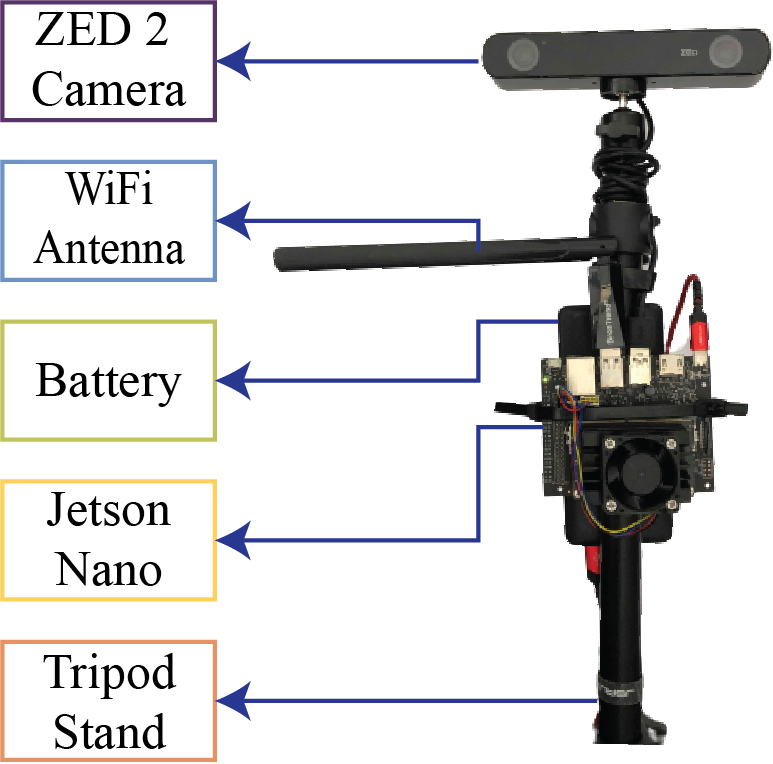}
    \caption {}
    \label{fig:ball-detection-module}
     \end{subfigure}
     \hfill
     \begin{subfigure}[b]{0.475\linewidth}
    \centering
    \resizebox{\linewidth}{4cm}{
\begin{tikzpicture}

\definecolor{cornflowerblue119170221}{RGB}{119,170,221}
\definecolor{crimson230928}{RGB}{230,9,28}
\definecolor{darkgray176}{RGB}{176,176,176}
\definecolor{royalblue1373251}{RGB}{13,73,251}

\begin{axis}[
legend cell align={left},
legend style={
  fill opacity=0.8,
  draw opacity=1,
  text opacity=1,
  at={(0.03,0.97)},
  anchor=north west,
  draw=none
},
tick pos=both,
x grid style={darkgray176},
xlabel={\large Ground Truth Depth (\(\displaystyle m\))},
xmin=2.675, xmax=10.925,
xtick style={color=black},
y grid style={darkgray176},
ylabel={\large Measured Depth (\(\displaystyle m\))},
ymin=2.505, ymax=12.163,
ytick style={color=black}
]
\path [draw=cornflowerblue119170221, fill=cornflowerblue119170221]
(axis cs:3.05,3.046)
--(axis cs:3.05,2.944)
--(axis cs:3.55,3.418)
--(axis cs:4.05,3.925)
--(axis cs:4.55,4.46)
--(axis cs:5.05,5.007)
--(axis cs:5.55,5.396)
--(axis cs:6.05,6.002)
--(axis cs:6.55,6.536)
--(axis cs:7.05,7.003)
--(axis cs:7.55,7.559)
--(axis cs:8.05,7.962)
--(axis cs:8.55,8.641)
--(axis cs:9.05,9.112)
--(axis cs:9.55,9.497)
--(axis cs:10.05,10.24)
--(axis cs:10.55,10.496)
--(axis cs:10.55,11.724)
--(axis cs:10.55,11.724)
--(axis cs:10.05,11.28)
--(axis cs:9.55,10.369)
--(axis cs:9.05,9.64)
--(axis cs:8.55,9.319)
--(axis cs:8.05,8.902)
--(axis cs:7.55,8.115)
--(axis cs:7.05,7.441)
--(axis cs:6.55,6.81)
--(axis cs:6.05,6.262)
--(axis cs:5.55,5.82)
--(axis cs:5.05,5.183)
--(axis cs:4.55,4.64)
--(axis cs:4.05,4.129)
--(axis cs:3.55,3.544)
--(axis cs:3.05,3.046)
--cycle;

\addplot [royalblue1373251]
table {%
3.05 2.995
3.55 3.481
4.05 4.027
4.55 4.55
5.05 5.095
5.55 5.608
6.05 6.132
6.55 6.673
7.05 7.222
7.55 7.837
8.05 8.432
8.55 8.98
9.05 9.376
9.55 9.933
10.05 10.76
10.55 11.11
};
\addlegendentry{\large Measured Depth}
\addplot [crimson230928, dashed]
table {%
3.05 3.05
3.55 3.55
4.05 4.05
4.55 4.55
5.05 5.05
5.55 5.55
6.05 6.05
6.55 6.55
7.05 7.05
7.55 7.55
8.05 8.05
8.55 8.55
9.05 9.05
9.55 9.55
10.05 10.05
10.55 10.55
};
\addlegendentry{\large Expected Depth}
\end{axis}

\end{tikzpicture}
    }
    \caption{}
    \label{fig:ball_detect_variance}
     \end{subfigure}
     \vspace{2pt}
        \caption{Fig.~\ref{fig:ball-detection-module} depicts the ball detection rig, and Fig.~\ref{fig:ball_detect_variance} depicts the measured vs. actual distance of a ball.}
        \label{fig:combined}
\end{figure}

To track a tennis ball as it moves across the court, we built six ball detection rigs (Fig.~\ref{fig:ball-detection-module}). Each rig consists of a Stereolabs ZED2 stereo camera, which is connected to an NVIDIA Jetson Nano that communicates to a central onboard computer over Wi-Fi. Each rig is mounted on a \SI{4}{\meter} long tripod stand, is placed to maximize coverage of the court (Fig.~\ref{fig:court-setup}), and is calibrated using an AprilTag~\cite{apriltag2016} calibration box.

We then employ multiple computer vision techniques to detect an airborne tennis ball: color thresholding, background subtraction, and noise removal to locate the pixel center of the largest moving colored tennis ball. Orange tennis balls are used since orange provides a strong contrast against the green/blue tennis courts. The ball coordinates are computed relative to the ball detection rig's reference frame by applying epipolar geometry on the pixel centers from the rig. The ball's pose is then mapped to the world frame. The positional error covariance is modeled through a quadratic relationship (i.e., $aD^2 + bD + c$) where $D$ is the ball’s distance to the camera and $a$, $b$, and $c$ are tuned via experiments of ball rolling along a linear rail (Fig.~\ref{fig:ball_detect_variance}). A quadratic expression is utilized as it provides a simple and accurate approximation of the measurement error. By pursuing a distributed vision approach, image data is efficiently processed locally on each Jetson Nano and sent to the onboard computer via Wi-Fi. These detection modules are able to process the stereo camera’s 1080p images, produce a positional estimate, and transmit it 
to the central computer within \SI{100}{\milli\second} at a frequency of \SI{25}{\hertz}. By pooling detections from six cameras, the vision system is able to achieve up to a maximum of 150 estimates per second. 

\paragraph{Ball EFK and Roll-out Prediction}
\label{paragraph:ball-ekf-rollout}

The detection rigs’ ball position estimates are fused through a continuous-time Extended Kalman Filter (EKF)~\cite{sorenson1985kalman} to produce a single pose estimate. The EKF is based on the {\sffamily robot\_localization} package~\cite{MooreStouchKeneralizedEkf2014}, which is augmented to incorporate ballistic trajectory, inelastic bouncing, court friction, and ball-air interactions~\cite{airballinteractions} into the state-estimate prediction. Our localization approach allows the ball detection errors to be merged into a single covariance estimate, informing how much confidence should be placed on the ball's predicted trajectory. To enhance the EKF predictions, the state estimation is reset when a tennis ball is first detected, and any measurement delays are handled by reverting the EKF to a specified lag time and re-applying all measurements to the present. By rolling out the EKF’s state predictions forward in time, it is possible to estimate the tennis ball’s future trajectory on the court, enabling the wheelchair to move to the appropriate position. If the confidence in the predicted trajectory is high, ESTHER acts upon the prediction. Otherwise, the trajectory estimate is ignored until the EKF converges to higher confidence. The EKF and rollout trajectory estimator run at \SI{100}{\hertz} to ensure the high-level planners can incorporate timely estimates. 

\indent \indent Our decentralized, modular approach to ball state estimation and trajectory prediction enables us to create a low-cost vision system that can be quickly set up on any tennis court. While we can detect the ball up to a maximum of 150 times a second by pooling detections from six camera rigs (Figure \ref{fig:ball-detection-module}), unfortunately, this rate is not enough to infer precisely the spin on the tennis ball (i.e., Magnus effect forces). In future work, we aim to infer this effect by incorporating indirect estimation methods such as Adaptive EKF (AEKF)~\cite{AdaptiveEKF_mobilerobot}, and trajectory fitting~\cite{jonas-tabletenniskukaspin2020} into our EKF.

\paragraph{Wheelchair Localization}
\label{paragraph:wheelchair-localization}

The motion of ESTHER's wheelchair base can be modeled as a differential drive base that is equipped with three different sensors to determine the robot's state in the world as displayed in Fig.~\ref{fig:architecture}. 8192 Counts Per Revolution motor encoders provide the velocity and position of each wheel at \SI{250}{\hertz}. A ZED2 Inertial Measurement Unit (IMU) is used to obtain the linear velocity, angular velocity, and orientation of the motion base at \SI{400}{\hertz}. A Velodyne Puck LiDAR sensor is used to obtain an egocentric \SI{360}{\degree} point cloud with a \SI{30}{\degree} vertical field of view at \SI{20}{\hertz}. For localizing the wheelchair, we first create a map of the environment offline by recording IMU data and 3D point cloud data while manually driving the wheelchair around at slow speeds. We use the {\sffamily hdl\_graph\_slam}~\cite{hdl_localization} package to create a Point Cloud Data map from this recorded data for use online. Afterward, we use a point cloud scan matching algorithm~\cite{hdl_localization} to obtain an odometry estimate based on LIDAR and IMU readings. A differential drive controller gives a second odometry estimate using encoder data from wheels and IMU readings. Due to the dynamic nature of the task, the wheelchair experiences slip during periods of high acceleration and deceleration when starting from rest or braking hard, making the use of IMU essential for accurate state estimation. We follow  guidance from~\cite{MooreStouchKeneralizedEkf2014} for fusing these estimates to get the current state of the wheelchair.

\subsubsection{Planning}
\label{subsubsec:planning}

Given the predicted ball trajectory and current position of the wheelchair, the ``strategizer'', a behavior orchestrator, selects the desired interception point. The interception point is the point at which the ball crosses the hit plane that passes through the center of the robot. Based on this interception point, the strategizer determines the required wheelchair pose, approximate stroke trajectory, and stroke timing. The strategizer also ensures that the interception point is not too close to the ground and that the corresponding joint positions are within limits.

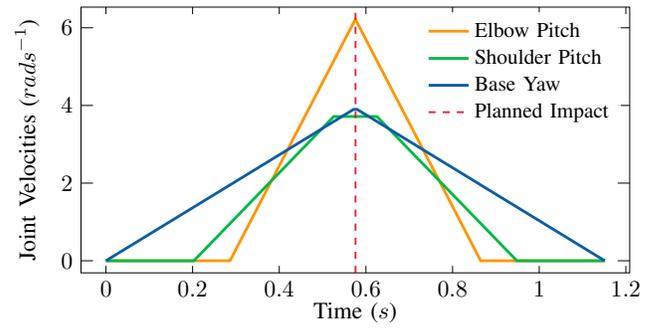
\begin{figure}
    \centering
    \resizebox{\linewidth}{!}{
\begin{tikzpicture}

\definecolor{crimson230928}{RGB}{230,9,28}
\definecolor{darkgray176}{RGB}{176,176,176}
\definecolor{darkorange2551490}{RGB}{255,149,0}
\definecolor{limegreen018569}{RGB}{0,185,69}
\definecolor{teal1293165}{RGB}{12,93,165}

\begin{axis}[
legend cell align={left},
legend style={fill opacity=0.8, draw opacity=1, text opacity=1, at={(0.82,1.4)}, draw=none},
tick pos=both,
x grid style={darkgray176},
xlabel={Time (\(\displaystyle s\))},
xmin=-0.05756673435, xmax=1.20890142135,
xtick style={color=black},
y grid style={darkgray176},
ylabel={Joint Velocities (\(\displaystyle rad s^{-1}\))},
ymin=-0.311046620299916, ymax=6.53197902629823,
yscale=0.7,
xscale=1.2,
ytick style={color=black}
]
\addplot [very thick, darkorange2551490]
table {%
0 0
0.202837876 0
0.28632165 0
0.525867017 5.15022539828014
0.571847507 6.13880593032345
0.575667343 6.22093240599832
0.579487179 6.13880593032345
0.625467669 5.15022539828014
0.865013037 0
0.948496811 0
1.151334687 0
};
\addlegendentry{\small Elbow Pitch}
\addplot [very thick, limegreen018569]
table {%
0 0
0.202837876 0
0.28632165 0.960063399539873
0.525867017 3.71483512420134
0.571847507 3.71483512420134
0.575667343 3.71483512420134
0.579487179 3.71483512420134
0.625467669 3.71483512420134
0.865013037 0.960063399539873
0.948496811 0
1.151334687 0
};
\addlegendentry{\small Shoulder Pitch}

\addplot [very thick, teal1293165]
table {%
0 0
0.202837876 1.38335431591587
0.28632165 1.95271365372996
0.525867017 3.58641305913789
0.571847507 3.9
0.575667343 3.9
0.579487179 3.9
0.625467669 3.58641305913789
0.865013037 1.95271365372996
0.948496811 1.38335431591587
1.151334687 0
};
\addlegendentry{\small Base Yaw}

\addplot [thick, crimson230928, opacity=0.9, dashed]
table {%
0.576 -0.311046620299916
0.576 6.53197902629823
};
\addlegendentry{\small Planned Impact}
\end{axis}

\end{tikzpicture}
    }
    \vspace{-10pt}
    \caption {Trapezoidal velocity profiles for individual joints.}
        \vspace{-10pt}
    \label{fig:trapezoidal-profile}
\end{figure}

Given the interception point, the strategizer finds the stroke parameters and the corresponding wheelchair placement for that stroke. The stroke is parameterized by three points: a start point (i.e., the joints at the swing's beginning), a contact point (i.e., the joints at ball contact), and an end point (i.e., the joints at the swing's completion). These points are determined such that the arm is fully extended at contact point while each joint is traveling at the maximum possible speed and staying within each individual joint's position, velocity, and acceleration limits. Using the contact point and the robot's kinematics, the wheelchair's desired position is geometrically determined. Lastly, the strategizer identifies the exact time to trigger the stroke by combining the stroke duration and the time the ball takes to reach the interception point. 

The strategizer continuously updates the interception point as the vision system updates its estimate of the ball's trajectory, allowing us to adjust  stroke parameters and wheelchair position until a few milliseconds before the ball crosses the interception plane, increasing the chances of a successful hit. 

\subsubsection{Low-Level Control/Execution}
\label{subsubsec:control}
We next describe the interaction between the strategizer and the low-level controllers.

\paragraph{Wheelchair Control}
\label{paragraph:wheelchair-control}

Given the robot’s position and state in the world, we use ROS’s {\sffamily move\_base} package \cite{ros-guide} to command the robot to go to the desired position. We use the {\sffamily move\_base's} default global planner (i.e., Dijkstra's algorithm) as our global planner and the Timed Elastic Band (TEB) planner \cite{teb_1} as our local planner. The global planner finds a path between the wheelchair's current pose $p_{curr}$ and the desired pose $p_{dest}$ given a map and obstacles detected by the LiDAR. The local planner tries to follow this path as closely as possible while performing real-time collision checking and obstacle avoidance based on the continuous point cloud data stream received from the LiDAR. The plan from the local planner is translated to wheel velocities through a differential drive controller. The velocities are then communicated from the onboard computer to the microcontroller over serial and are executed on the wheelchair. The motor controller uses PID control for the wheels.

\begin{figure}
    \centering    \includegraphics[width=0.9\linewidth,height=3.3cm]{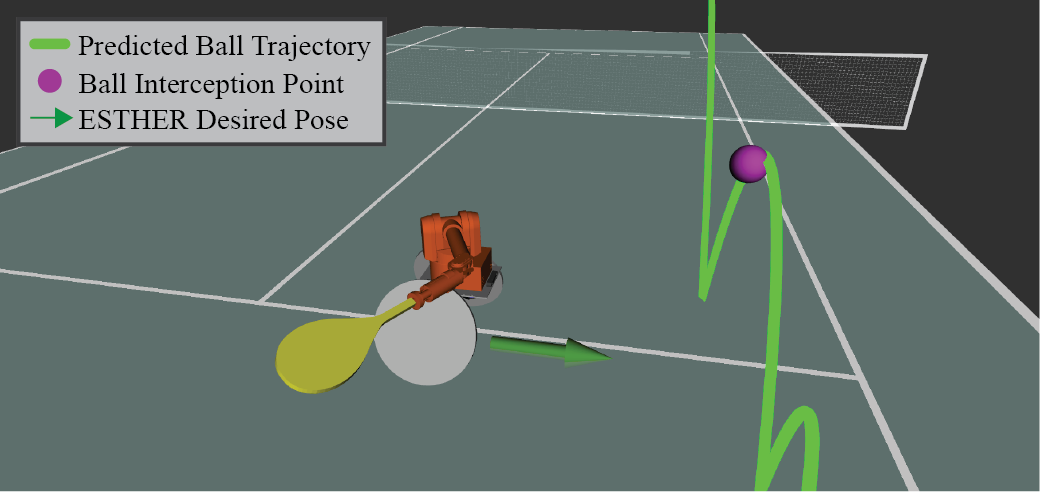}
    \caption {ESTHER simulation in RViz. }
    \label{fig:simulation-rviz}
\end{figure}

\paragraph{Arm Control}
\label{paragraph:arm-control}

The arm-control subsystem generates and executes joint trajectories. The Barrett high-speed WAM has 7 joints in total: base yaw, shoulder pitch, shoulder yaw, elbow pitch, wrist pitch, wrist yaw, and palm yaw. The swing utilizes base yaw, shoulder pitch, and elbow pitch joints to generate speed, and the rest of the joints are held at constant positions during the swing so the racket is in the right orientation when making contact with the ball.

\indent \indent Humans generate high racket head speeds by merging contributions from multiple joints~\cite{groundstroke_biodynamics}. Inspired by this ``summation of speed principle''~\cite{Landlinger2010-iy}, we created a ``Fully-Extended'' Ground stroke (FEG) to maximize  racket speed when making contact with the ball. The base yaw and elbow pitch joints provide the velocity component in the direction we want to return the ball, and the shoulder pitch joint adjusts for ball height and hits the ball upwards (Fig.~\ref{fig:swing-capability}). These joints follow a trapezoidal velocity profile (Fig.~\ref{fig:trapezoidal-profile}). Joint trajectories are sent from the onboard computer to the WAM computer, which uses a PID controller to execute the trajectories.

\subsection{Simulation}
\label{subsec:simulation}

We developed a kinematic simulator in RViz \cite{rviz} (Fig.~\ref{fig:simulation-rviz}) for testing the system for development efficiency and safety.

\section{Experiments and Results}
\label{sec:exp-and-results}

We evaluate ESTHER and its components to serve as a baseline for future research on athletic robots for tennis. In Section~\ref{subsec:exp_setup}, we describe the details of our experimental setup. In Section~\ref{subsec:system-capabilities}, we evaluate the capabilities of individual subsystems and the system as a whole. Finally, we report the results of our experiments in Section~\ref{subsec:court-lab-experiments}. 

\begin{figure}
    \centering
    \vspace{-2mm}
    \includegraphics[height = 5.7cm]{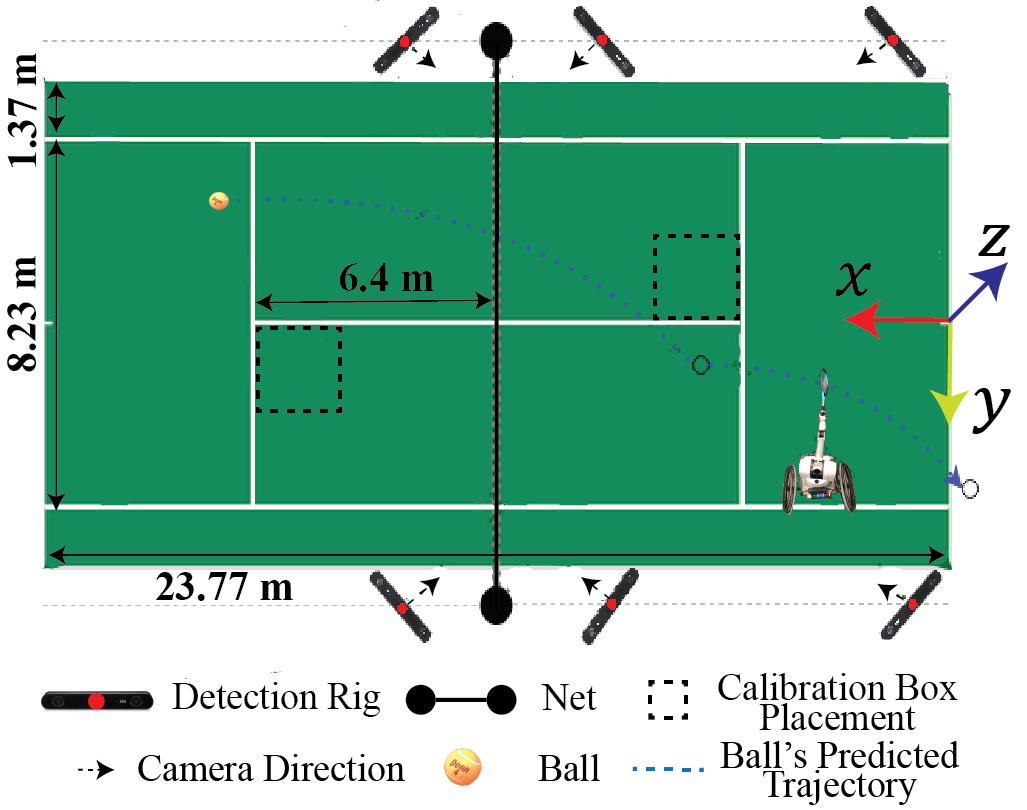}
    \caption {Schematic showing the court setup }
    \label{fig:court-setup}
\end{figure}

\subsection{Experimental Setup}
\label{subsec:exp_setup}

We have two evaluation settings: an NCAA Division 1 indoor tennis facility and within a lab space. The tennis court setup is a regulation court i.e. \SI{23.77}{\meter} by \SI{10.97}{\meter}, and the lab setup is \SI{11}{\meter} by \SI{4.57}{\meter}.
Fig.~\ref{fig:court-setup} displays the world coordinate frame and on-court setup. The ball is launched by a ball launcher 
or a human towards the robot from the other side of the court. The lab setup is similar to the court setup except the lab covers a smaller area, has a different surface texture affecting ball bounce and wheel traction, does not have court lines, and has different lighting conditions. Overall, the lab setup is easier for hitting balls.

\subsection{Sub-System Experiments}
\label{subsec:system-capabilities}

In this section, we overview the capabilities of our current system and the results of experiments conducted in the tennis court and lab based on the setup described in Section~\ref{subsec:exp_setup}.

\subsubsection{Wheelchair}
\label{subsec:wheelchair-capability}

Manual Mode -- We safely drove the wheelchair with load at a linear speed of \SI{4.34}{\meter\per\second} and at an angular speed of \SI{5.8}{\radian\per\second}, less than half of the maximum possible speeds achievable with our system.

Autonomous Mode -- We achieve accelerations of \SI{1.42}{\meter\second\tothe{-2}} and deceleration of \SI{1.60}{\meter\second\tothe{-2}} which is more than the average side-to-side acceleration and deceleration values of $\approx$\SI{1.00}{\meter\second\tothe{-2}} achieved by professional human players~\cite{Filipcic2017-iy}.

\subsubsection{Stroke Speed}
\label{subsec:arm-swing-capability}

\begin{figure}
    \centering
    \resizebox{\linewidth}{!}{
        \input{tikz/racket_speed.tex}
    }
    \includegraphics[width=0.9\linewidth]{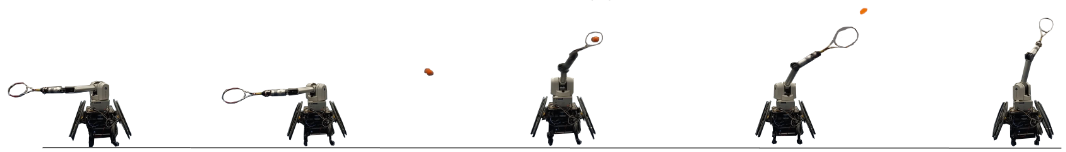}
    \caption {Racket head speed during a FEG stroke.}
    \label{fig:swing-capability}
\end{figure}

ESTHER reaches pre-impact racket head speeds (Fig.~\ref{fig:swing-capability}) of \SI{10}{\meter\per\second}. This speed is on the same order of magnitude of professional players (\SI{17}-\SI{36}{\meter\per\second} \cite{Allen2016}).

\subsubsection{Ball Rollout Prediction}
\label{subsec:sensing-state-estimation-capabilities}

\begin{figure}
    \centering
    \includegraphics[width=\linewidth, height = 4.36cm]{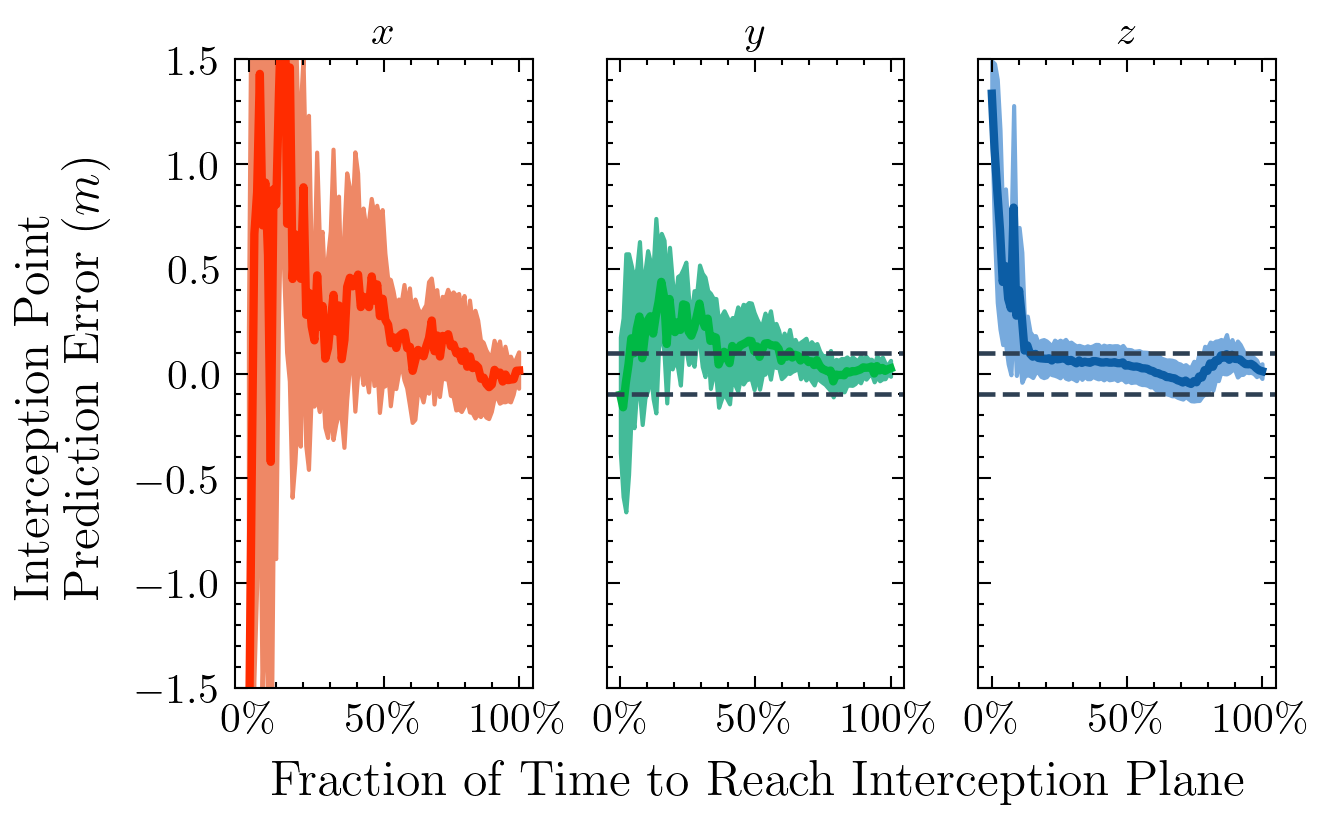}
    \caption{Error in the ball's forecasted vs. measured position at the desired interception plane. The bold lines indicate the mean error, and the shaded region represents one standard deviation of the errors across 10 trajectories. The dashed lines represent the range of acceptable error, defined to be equal to the racket head width ($\approx$ \SI{20}{\cm})}.
    \label{fig:rollout_error}
    \vspace{-4mm}
\end{figure}

An important consideration for playing a sport like tennis is to anticipate the ball trajectory early so the robot can get in position to return the ball. To measure the performance of our ball trajectory prediction system, we measure the error between the rollout's predicted interception point and the observed interception point that is obtained after the entire ball trajectory is observed. In Fig.~\ref{fig:rollout_error} we visualize the interception point prediction error as a function of the fraction of the total time taken to reach the interception plane for 10 ball trajectories in the court setup. The ball trajectories vary slightly in the total flight time but are roughly \SI{2}{\second} long, and the balls were launched at roughly \SI{8}{\meter\per\second} from the ball launcher. In the court setup, the ball travels along the \(x\)-axis, thereby the prediction error in \(x\) affects the timing of the stroke. Similarly, prediction error in \(y\) affects wheelchair positioning, and \(z\) affects the height at which the arm swings. As the FEG stroke takes about \SI{0.5}{\second} to go from the start to the interception point, we must be certain about the \(x\) coordinate of the interception point before 75\% of the total trajectory has elapsed. Defining acceptable error margins for racket positions at the interception plane to be equal to the racket head width, we can observe from  Fig.~\ref{fig:rollout_error} that the \(y\) and \(z\) predictions converge to be within the acceptable error margin after 50\% of the trajectory is seen. Given that the wheelchair can accelerate at \SI{1.42}{\meter\second\tothe{-2}}, we assess that ESTHER is capable of moving up to \SI{0.71}{\meter} from a standing start after we establish confidence in the predicted interception point. For larger distances, we need more accurate predictions quicker.

\subsubsection{Reachability}
\label{subsec:reachability}

Researchers found that in professional tennis, players typically move less than \SI{2}{\meter} for the majority of the strokes~\cite{tennis-lateral-movement}. To benchmark our system's ability to reach balls in this range, we launched balls toward the robot in our court setup and recorded the the distance required by the wheelchair to move to intercept the ball. After every 10 balls, we increased the distance between the wheelchair and the ball's average position at the interception plane by \SI{0.3}{\meter}. We performed four sets of such trials for a total of 40 trials. Due to variance in the ball launcher, after 40 ball launches, we had observed the required movement to intercept the ball had reached 2 m which was comparable to the typical distance moved by human players between strokes. In this test, to mitigate the latency introduced by the time it requires the vision system to converge to an accurate estimate, we start moving the wheelchair towards the ball's average detected \(y\) position as soon as the ball is detected for the first time after launch. The position of the wheelchair is later fine-tuned once the prediction from the vision system has converged. Fig.~\ref{fig:system-reachibility} depicts a histogram of the success rate as a function of the distance the wheelchair needed to move.  

\subsection{Whole-System Experiments}
\label{subsec:court-lab-experiments}

\begin{figure}
    \centering
    \includegraphics{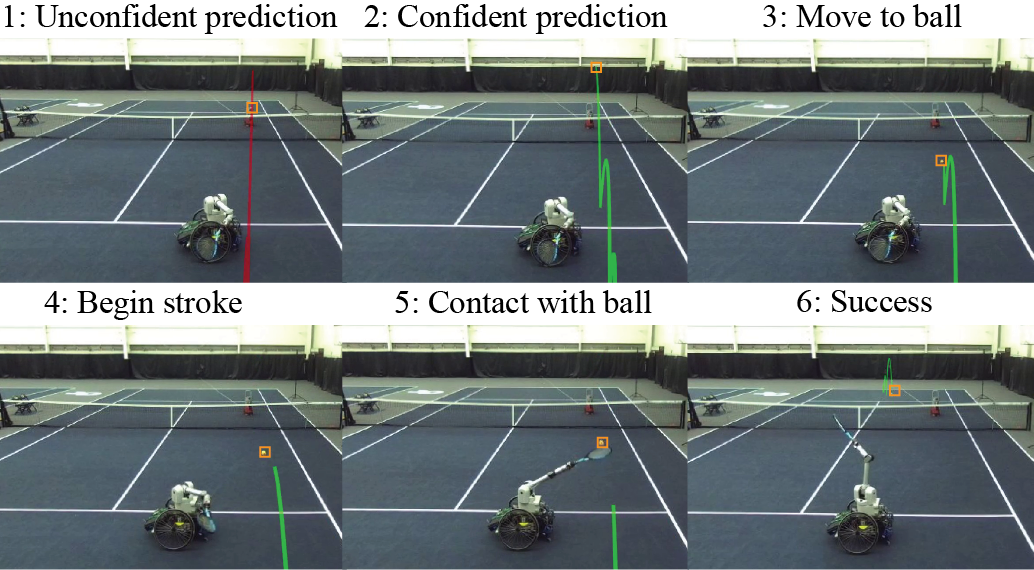}
    \caption{ESTHER in action. Orange box highlights ball.}
    \label{fig:frame-by-frame}
\end{figure}

We perform 15 trials in each scenario and report the details of the hit and return rate in Table~\ref{tab:results}. A returned ball over the net landing inside the singles lines of the court is marked as successful. Fig.~\ref{fig:frame-by-frame} illustrates a frame-by-frame depiction of our system in action on the court. For tests inside the lab, a ball that goes over the net height (\SI{1.07}{\meter}) while crossing the plane from which the ball was launched is marked successful.

\begin{table*}[ht]
\begin{center}
\begin{tabular}{lcccccc}
\hline
  & 
  \multicolumn{2}{c}{\bf Ball Launch} & \multicolumn{2}{c}{\bf Interception Point Lateral ($y$) spread} & 
  \multicolumn{2}{c}{\bf Success Metrics}
  \\ \cline{2-7}
  {\bf Setup} & {\bf \Centerstack{Distance to \\ Interception Plane (\SI{}{\meter})}} & {\bf \Centerstack{Average Launch\\Speed (\SI{}{\meter\per\second})}} & {\bf IQR (\SI{}{\meter})} & {\bf Std. Dev (\SI{}{\meter})} & {\bf Hit Rate} & {\bf Success Rate} \\
  \hline
    \shortstack{Court (Ball Launcher)} & 7.9 & 8.01 & 0.31 & 0.23 & 73\% & 66\%  \\
    \shortstack{Court (Ball Launcher)} & 12.8 & 12.64 & 0.26 & 0.29 & 60\% & 53\% \\ 
    \shortstack{Lab (Ball Launcher)} & 7.5 & 6.79 & 0.28 & 0.20 & 93\% & 80\% \\
    \shortstack{Lab (Human )} & 7.5 & 6.56 & 0.54 & 0.52 & 40\% & 33\% \\ \hline
    \end{tabular}
\caption{Experiment results for 15 consecutive trials in different scenarios}
\label{tab:results}
\end{center}
\end{table*}

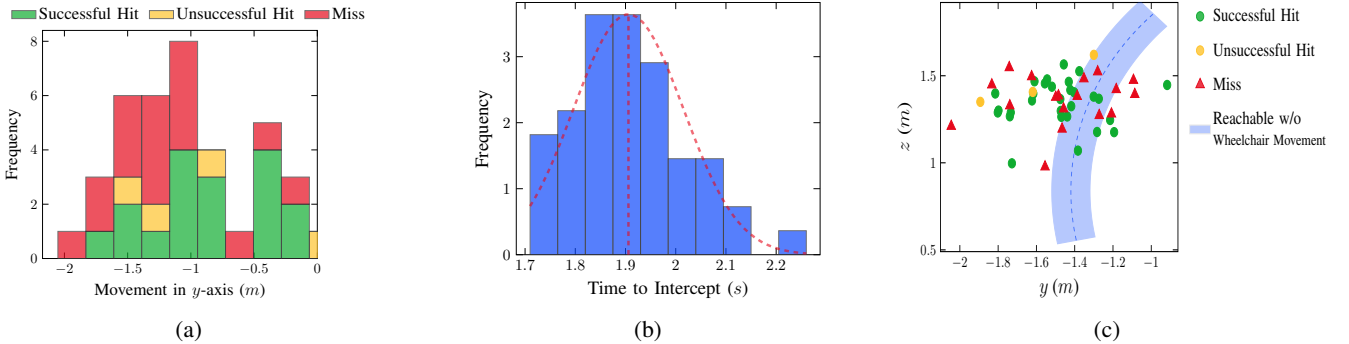
\begin{figure*}
     \centering
     \begin{subfigure}[b]{0.32\linewidth}
    \centering
    \resizebox{!}{4cm}{
\begin{tikzpicture}

\definecolor{crimson230928}{RGB}{230,9,28}
\definecolor{darkgray176}{RGB}{176,176,176}
\definecolor{darkslategray51}{RGB}{51,51,51}
\definecolor{forestgreen1517248}{RGB}{15,172,48}
\definecolor{goldenrod25419545}{RGB}{254,195,45}

\begin{axis}[
legend cell align={left},
legend columns=-1,
legend style={fill opacity=0.8, draw opacity=1, text opacity=1, at={(1.2,1.125)}, draw=none}, 
tick pos=both,
x grid style={darkgray176},
xlabel={\large Movement in \(\displaystyle y\)-axis (\(\displaystyle m\))},
xmin=-2.17304273250087, xmax=0,
xtick style={color=black},
y grid style={darkgray176},
ylabel={\large Frequency},
ymin=0, ymax=8.4,
ytick style={color=black}
]
\addlegendimage{area legend,draw=darkslategray51,fill=forestgreen1517248,opacity=0.7}
\addlegendentry{\large Successful Hit\:\:}

\draw[draw=darkslategray51,fill=forestgreen1517248,opacity=0.7] (axis cs:-2.05151581075124,0) rectangle (axis cs:-1.83055777120645,0);
\draw[draw=darkslategray51,fill=forestgreen1517248,opacity=0.7] (axis cs:-1.83055777120645,0) rectangle (axis cs:-1.60959973166167,1);
\draw[draw=darkslategray51,fill=forestgreen1517248,opacity=0.7] (axis cs:-1.60959973166167,0) rectangle (axis cs:-1.38864169211689,2);
\draw[draw=darkslategray51,fill=forestgreen1517248,opacity=0.7] (axis cs:-1.38864169211689,0) rectangle (axis cs:-1.1676836525721,1);
\draw[draw=darkslategray51,fill=forestgreen1517248,opacity=0.7] (axis cs:-1.1676836525721,0) rectangle (axis cs:-0.94672561302732,4);
\draw[draw=darkslategray51,fill=forestgreen1517248,opacity=0.7] (axis cs:-0.94672561302732,0) rectangle (axis cs:-0.725767573482537,3);
\draw[draw=darkslategray51,fill=forestgreen1517248,opacity=0.7] (axis cs:-0.725767573482537,0) rectangle (axis cs:-0.504809533937753,0);
\draw[draw=darkslategray51,fill=forestgreen1517248,opacity=0.7] (axis cs:-0.504809533937753,0) rectangle (axis cs:-0.283851494392969,4);
\draw[draw=darkslategray51,fill=forestgreen1517248,opacity=0.7] (axis cs:-0.283851494392969,0) rectangle (axis cs:-0.0628934548481856,2);
\draw[draw=darkslategray51,fill=forestgreen1517248,opacity=0.7] (axis cs:-0.0628934548481856,0) rectangle (axis cs:0.158064584696598,0);
\draw[draw=darkslategray51,fill=forestgreen1517248,opacity=0.7] (axis cs:0.158064584696598,0) rectangle (axis cs:0.379022624241381,1);

\addlegendimage{area legend,draw=darkslategray51,fill=goldenrod25419545,opacity=0.7}
\addlegendentry{\large Unsuccessful Hit\:\:}

\draw[draw=darkslategray51,fill=goldenrod25419545,opacity=0.7] (axis cs:-2.05151581075124,0) rectangle (axis cs:-1.83055777120645,0);
\draw[draw=darkslategray51,fill=goldenrod25419545,opacity=0.7] (axis cs:-1.83055777120645,1) rectangle (axis cs:-1.60959973166167,1);
\draw[draw=darkslategray51,fill=goldenrod25419545,opacity=0.7] (axis cs:-1.60959973166167,2) rectangle (axis cs:-1.38864169211689,3);
\draw[draw=darkslategray51,fill=goldenrod25419545,opacity=0.7] (axis cs:-1.38864169211689,1) rectangle (axis cs:-1.1676836525721,2);
\draw[draw=darkslategray51,fill=goldenrod25419545,opacity=0.7] (axis cs:-1.1676836525721,4) rectangle (axis cs:-0.94672561302732,4);
\draw[draw=darkslategray51,fill=goldenrod25419545,opacity=0.7] (axis cs:-0.94672561302732,3) rectangle (axis cs:-0.725767573482537,4);
\draw[draw=darkslategray51,fill=goldenrod25419545,opacity=0.7] (axis cs:-0.725767573482537,0) rectangle (axis cs:-0.504809533937753,0);
\draw[draw=darkslategray51,fill=goldenrod25419545,opacity=0.7] (axis cs:-0.504809533937753,4) rectangle (axis cs:-0.283851494392969,4);
\draw[draw=darkslategray51,fill=goldenrod25419545,opacity=0.7] (axis cs:-0.283851494392969,2) rectangle (axis cs:-0.0628934548481856,2);
\draw[draw=darkslategray51,fill=goldenrod25419545,opacity=0.7] (axis cs:-0.0628934548481856,0) rectangle (axis cs:0.158064584696598,1);
\draw[draw=darkslategray51,fill=goldenrod25419545,opacity=0.7] (axis cs:0.158064584696598,1) rectangle (axis cs:0.379022624241381,1);

\addlegendimage{area legend,draw=darkslategray51,fill=crimson230928,opacity=0.7}
\addlegendentry{\large Miss}

\draw[draw=darkslategray51,fill=crimson230928,opacity=0.7] (axis cs:-2.05151581075124,0) rectangle (axis cs:-1.83055777120645,1);
\draw[draw=darkslategray51,fill=crimson230928,opacity=0.7] (axis cs:-1.83055777120645,1) rectangle (axis cs:-1.60959973166167,3);
\draw[draw=darkslategray51,fill=crimson230928,opacity=0.7] (axis cs:-1.60959973166167,3) rectangle (axis cs:-1.38864169211689,6);
\draw[draw=darkslategray51,fill=crimson230928,opacity=0.7] (axis cs:-1.38864169211689,2) rectangle (axis cs:-1.1676836525721,6);
\draw[draw=darkslategray51,fill=crimson230928,opacity=0.7] (axis cs:-1.1676836525721,4) rectangle (axis cs:-0.94672561302732,8);
\draw[draw=darkslategray51,fill=crimson230928,opacity=0.7] (axis cs:-0.94672561302732,4) rectangle (axis cs:-0.725767573482537,4);
\draw[draw=darkslategray51,fill=crimson230928,opacity=0.7] (axis cs:-0.725767573482537,0) rectangle (axis cs:-0.504809533937753,1);
\draw[draw=darkslategray51,fill=crimson230928,opacity=0.7] (axis cs:-0.504809533937753,4) rectangle (axis cs:-0.283851494392969,5);
\draw[draw=darkslategray51,fill=crimson230928,opacity=0.7] (axis cs:-0.283851494392969,2) rectangle (axis cs:-0.0628934548481856,3);
\draw[draw=darkslategray51,fill=crimson230928,opacity=0.7] (axis cs:-0.0628934548481856,1) rectangle (axis cs:0.158064584696598,1);
\draw[draw=darkslategray51,fill=crimson230928,opacity=0.7] (axis cs:0.158064584696598,1) rectangle (axis cs:0.379022624241381,2);
\end{axis}

\end{tikzpicture}
    }
    \caption{}
    \label{fig:system-reachibility}
     \end{subfigure}
     \hfill
     \begin{subfigure}[b]{0.32\linewidth}
    \centering
    \resizebox{!}{4cm}{
\begin{tikzpicture}

\definecolor{crimson230928}{RGB}{230,9,28}
\definecolor{darkgray176}{RGB}{176,176,176}
\definecolor{darkslategray51}{RGB}{51,51,51}
\definecolor{royalblue1373251}{RGB}{13,73,251}

\begin{axis}[
tick pos=both,
x grid style={darkgray176},
xlabel={\large Time to Intercept (\(\displaystyle s\))},
xmin=1.6825, xmax=2.2875,
xtick style={color=black},
y grid style={darkgray176},
ylabel={\large Frequency},
ymin=0, ymax=3.82104870629353,
ytick style={color=black}
]
\draw[draw=darkslategray51,fill=royalblue1373251,opacity=0.7] (axis cs:1.71,0) rectangle (axis cs:1.765,1.81818181818182);
\draw[draw=darkslategray51,fill=royalblue1373251,opacity=0.7] (axis cs:1.765,0) rectangle (axis cs:1.82,2.18181818181818);
\draw[draw=darkslategray51,fill=royalblue1373251,opacity=0.7] (axis cs:1.82,0) rectangle (axis cs:1.875,3.63636363636363);
\draw[draw=darkslategray51,fill=royalblue1373251,opacity=0.7] (axis cs:1.875,0) rectangle (axis cs:1.93,3.63636363636364);
\draw[draw=darkslategray51,fill=royalblue1373251,opacity=0.7] (axis cs:1.93,0) rectangle (axis cs:1.985,2.90909090909091);
\draw[draw=darkslategray51,fill=royalblue1373251,opacity=0.7] (axis cs:1.985,0) rectangle (axis cs:2.04,1.45454545454545);
\draw[draw=darkslategray51,fill=royalblue1373251,opacity=0.7] (axis cs:2.04,0) rectangle (axis cs:2.095,1.45454545454546);
\draw[draw=darkslategray51,fill=royalblue1373251,opacity=0.7] (axis cs:2.095,0) rectangle (axis cs:2.15,0.727272727272725);
\draw[draw=darkslategray51,fill=royalblue1373251,opacity=0.7] (axis cs:2.15,0) rectangle (axis cs:2.205,0);
\draw[draw=darkslategray51,fill=royalblue1373251,opacity=0.7] (axis cs:2.205,0) rectangle (axis cs:2.26,0.363636363636363);
\addplot [thick, crimson230928, opacity=0.6, dashed, style=ultra thick]
table {%
1.71 0.736217061782989
1.72122448979592 0.879687143436602
1.73244897959184 1.040131297189
1.74367346938776 1.21698615212584
1.75489795918367 1.40903136726097
1.76612244897959 1.61433342409631
1.77734693877551 1.83022030334047
1.78857142857143 2.05329359099195
1.79979591836735 2.27948262896629
1.81102040816327 2.50414270712208
1.82224489795918 2.72219615156276
1.8334693877551 2.92831174243805
1.84469387755102 3.11711449713827
1.85591836734694 3.28341481630256
1.86714285714286 3.42244363885749
1.87836734693878 3.53007887120572
1.88959183673469 3.60304814497861
1.90081632653061 3.63909400599384
1.91204081632653 3.6370899041743
1.92326530612245 3.59709867011979
1.93448979591837 3.52036924353942
1.94571428571429 3.40927189120491
1.9569387755102 3.26717660241657
1.96816326530612 3.09828336563735
1.97938775510204 2.90741624748643
1.99061224489796 2.699795340373
2.00183673469388 2.48080156083788
2.0130612244898 2.25574894094912
2.02428571428571 2.02967756116037
2.03551020408163 1.80717783537667
2.04673469387755 1.59225376702361
2.05795918367347 1.38822937957787
2.06918367346939 1.19769911925301
2.08040816326531 1.02251993027373
2.09163265306122 0.863840150785738
2.10285714285714 0.722158525606964
2.11408163265306 0.597405549047779
2.12530612244898 0.489039022154036
2.1365306122449 0.396146049341073
2.14775510204082 0.31754457430622
2.15897959183673 0.251878800645242
2.17020408163265 0.197704288270269
2.18142857142857 0.153560004004813
2.19265306122449 0.118026001091334
2.20387755102041 0.0897666097746205
2.21510204081633 0.0675599791318668
2.22632653061224 0.0503154942009786
2.23755102040816 0.0370810079940589
2.24877551020408 0.0270420039911415
2.26 0.0195147847999877
};
\addplot [crimson230928, opacity=0.6, dashed, style=ultra thick]
table {%
1.90584 0
1.90584 3.63909400599384
};
\end{axis}

\end{tikzpicture}
    }
    \caption{}
    \label{fig:intercept_time_hist}
     \end{subfigure}
     \hfill
     \begin{subfigure}[b]{0.32\linewidth}
         \centering
    \resizebox{6cm}{4cm}{
\begin{tikzpicture}

\definecolor{crimson230928}{RGB}{230,9,28}
\definecolor{darkgray176}{RGB}{176,176,176}
\definecolor{forestgreen1517248}{RGB}{15,172,48}
\definecolor{goldenrod25419545}{RGB}{254,195,45}
\definecolor{royalblue1373251}{RGB}{13,73,251}

\begin{axis}[
legend cell align={left},
legend style={fill opacity=0.8, draw opacity=1, text opacity=1, at={(1.65,1)}, draw=none, row sep=0.3cm, cells={align=left}},
tick pos=both,
x grid style={darkgray176},
xlabel={\Large \(\displaystyle y\) (\(\displaystyle m\))},
xmin=-2.1014, xmax=-0.8606,
xtick style={color=black},
y grid style={darkgray176},
ylabel={\Large \(\displaystyle z\) (\(\displaystyle m\))},
ymin=0.489851745330698, ymax=1.92068345401764,
ytick style={color=black}
]
\addplot [draw=forestgreen1517248, fill=forestgreen1517248, mark=*, only marks, mark size=3pt]
table{%
x  y
-1.556 1.456
-1.475 1.366
-1.439 1.266
-1.403 1.405
-1.815 1.398
-1.519 1.437
-1.798 1.3
-1.623 1.358
-1.739 1.267
-1.544 1.479
-1.423 1.418
-1.734 1.289
-1.215 1.246
-1.418 1.325
-1.616 1.397
-1.431 1.465
-1.61 1.468
-1.47 1.264
-1.473 1.298
-0.917 1.447
-1.376 1.526
-1.283 1.177
-1.301 1.38
-1.383 1.07
-1.194 1.176
-1.728 0.997
-1.457 1.565
-1.274 1.368
-1.802 1.289
};
\addlegendentry{\large Successful Hit}
\addplot [draw=goldenrod25419545, fill=goldenrod25419545, mark=*, only marks, mark size=3pt]
table{%
x  y
-1.893 1.35
-1.3 1.62
-1.619 1.407
};
\addlegendentry{\large Unsuccessful Hit}
\addplot [draw=crimson230928, fill=crimson230928, mark=triangle*, only marks, mark size=4pt]
table{%
x  y
-1.208 1.281
-1.272 1.272
-2.045 1.21
-1.555 0.977
-1.499 1.377
-1.625 1.495
-1.739 1.328
-1.485 1.385
-1.458 1.31
-1.183 1.421
-1.387 1.382
-1.28 1.523
-1.086 1.393
-1.742 1.547
-1.093 1.474
-1.466 1.194
-1.352 1.483
-1.833 1.448
};
\addlegendentry{\large Miss}

\addlegendimage{area legend,draw=darkgray176,fill=royalblue1373251,opacity=0.3}
\addlegendentry{{\large Reachable w/o \\[0.2em] Wheelchair Movement}}
\path [draw=royalblue1373251, fill=royalblue1373251, opacity=0.3]
(axis cs:-1.48970119831869,0.535022617191507)
--(axis cs:-1.49555335987538,0.565485455701816)
--(axis cs:-1.50078265535408,0.596061374229105)
--(axis cs:-1.50538690686473,0.626737638555413)
--(axis cs:-1.50936419683477,0.657501472670897)
--(axis cs:-1.51271286880778,0.688340064094762)
--(axis cs:-1.51543152813327,0.71924056921139)
--(axis cs:-1.51751904254764,0.750190118619429)
--(axis cs:-1.51897454264563,0.781175822491628)
--(axis cs:-1.51979742224252,0.81218477594317)
--(axis cs:-1.51998733862649,0.843204064406285)
--(axis cs:-1.51954421270142,0.874220769008892)
--(axis cs:-1.51846822901981,0.905221971955032)
--(axis cs:-1.51675983570592,0.936194761904856)
--(axis cs:-1.51441974426911,0.96712623935192)
--(axis cs:-1.51144892930758,0.998003521995559)
--(axis cs:-1.50784862810238,1.02881375010609)
--(axis cs:-1.50362034010219,1.0595440918806)
--(axis cs:-1.49876582629879,1.09018174878715)
--(axis cs:-1.49328710849362,1.12071396089505)
--(axis cs:-1.48718646845582,1.15112801218915)
--(axis cs:-1.48046644697185,1.18141123586573)
--(axis cs:-1.47312984278733,1.211551019608)
--(axis cs:-1.46517971144143,1.24153481083883)
--(axis cs:-1.45661936399431,1.27135012194865)
--(axis cs:-1.44745236564811,1.30098453549627)
--(axis cs:-1.43768253426213,1.33042570938048)
--(axis cs:-1.42731393876277,1.35966138198026)
--(axis cs:-1.41635089744892,1.3886793772615)
--(axis cs:-1.40479797619349,1.41746760984803)
--(axis cs:-1.39265998654176,1.44601409005498)
--(axis cs:-1.37994198370758,1.47430692888215)
--(axis cs:-1.36664926446786,1.5023343429656)
--(axis cs:-1.35278736495667,1.5300846594851)
--(axis cs:-1.33836205835953,1.55754632102568)
--(axis cs:-1.32337935250897,1.58470789039099)
--(axis cs:-1.30784548738243,1.61155805536667)
--(axis cs:-1.29176693250345,1.63808563343167)
--(axis cs:-1.27515038424722,1.66427957641548)
--(axis cs:-1.25800276305169,1.6901289750995)
--(axis cs:-1.24033121053537,1.71562306376048)
--(axis cs:-1.22214308652301,1.74075122465422)
--(axis cs:-1.20344596598034,1.76550299243763)
--(axis cs:-1.18424763585932,1.78986805852732)
--(axis cs:-1.16455609185497,1.8138362753929)
--(axis cs:-1.14437953507541,1.83739766078324)
--(axis cs:-1.12372636862622,1.86054240188383)
--(axis cs:-1.10260519411076,1.88326085940361)
--(axis cs:-1.08102480804773,1.90554357158955)
--(axis cs:-1.05899419820769,1.92738125816727)
--(axis cs:-0.919652856338258,1.78391003998737)
--(axis cs:-0.938784701725663,1.76494573322251)
--(axis cs:-0.95752556330671,1.7455949568505)
--(axis cs:-0.975867635912247,1.72586577005701)
--(axis cs:-0.993803280460224,1.70576638962755)
--(axis cs:-1.01132502713721,1.68530518652542)
--(axis cs:-1.0284255785094,1.6644906824053)
--(axis cs:-1.04509781256188,1.64333154606426)
--(axis cs:-1.06133478566472,1.6218365898313)
--(axis cs:-1.07712973546493,1.60001476589726)
--(axis cs:-1.09247608370278,1.57787516258641)
--(axis cs:-1.10736743895153,1.55542700057134)
--(axis cs:-1.12179759927931,1.53267962903276)
--(axis cs:-1.13576055483211,1.50964252176579)
--(axis cs:-1.14925049033674,1.48632527323428)
--(axis cs:-1.16226178752275,1.46273759457493)
--(axis cs:-1.17478902746237,1.43888930955285)
--(axis cs:-1.18682699282735,1.41479035047012)
--(axis cs:-1.19837067006184,1.39045075402924)
--(axis cs:-1.20941525147048,1.36588065715301)
--(axis cs:-1.21995613722066,1.34109029276277)
--(axis cs:-1.22998893725827,1.31608998551656)
--(axis cs:-1.23950947313609,1.29089014750917)
--(axis cs:-1.24851377975395,1.26550127393568)
--(axis cs:-1.2569981070102,1.23993393872045)
--(axis cs:-1.26495892136348,1.2141987901133)
--(axis cs:-1.2723929073044,1.18830654625477)
--(axis cs:-1.27929696873636,1.16226799071221)
--(axis cs:-1.28566823026502,1.13609396798866)
--(axis cs:-1.29150403839584,1.10979537900637)
--(axis cs:-1.2968019626392,1.08338317656676)
--(axis cs:-1.30155979652263,1.05686836078884)
--(axis cs:-1.3057755585098,1.03026197452789)
--(axis cs:-1.30944749282575,1.00357509877634)
--(axis cs:-1.31257407018816,0.976818848048775)
--(axis cs:-1.31515398844423,0.950004365752983)
--(axis cs:-1.31718617311303,0.923142819548953)
--(axis cs:-1.318669777833,0.896245396697791)
--(axis cs:-1.31960418471439,0.869323299402459)
--(axis cs:-1.31998900459669,0.8423877401423)
--(axis cs:-1.31982407721061,0.815449937003279)
--(axis cs:-1.31910947124489,0.788521109005887)
--(axis cs:-1.31784548431768,0.761612471432662)
--(axis cs:-1.31603264285258,0.73473523115726)
--(axis cs:-1.31367170185938,0.70790058197703)
--(axis cs:-1.31076364461967,0.681119699951042)
--(axis cs:-1.30730968227726,0.65440373874549)
--(axis cs:-1.3033112533338,0.627763824988433)
--(axis cs:-1.29877002304968,0.601211053635788)
--(axis cs:-1.29368788275044,0.574756483350519)
--cycle;

\addplot [royalblue1373251, opacity=0.7, dashed, forget plot]
table {%
-1.39169454053456 0.554889550271013
-1.39716169146253 0.583348254668802
-1.40204695434394 0.611912599608769
-1.40634829457099 0.640570688650452
-1.41006392072722 0.669310586310969
-1.41319228533358 0.698120323035896
-1.41573208549292 0.726987900184325
-1.41768226343266 0.755901295026046
-1.41904200694526 0.784848465748758
-1.41981074972656 0.813817356473224
-1.41998817161159 0.842795902274292
-1.41957419870791 0.871772034205675
-1.4185690034264 0.900733684326412
-1.41697300440947 0.929668790726905
-1.41478686635667 0.958565302552452
-1.41201149974787 0.987411185022167
-1.40864806046407 1.01619442444121
-1.404697949306 1.04490303320424
-1.40016281141071 1.07352505478799
-1.39504453556641 1.10204856873091
-1.38934525342583 1.13046169559776
-1.38306733861843 1.1587526019272
-1.37621340576184 1.1869095051601
-1.36878630937292 1.2149206785468
-1.3607891426789 1.24277445603097
-1.35222523632916 1.27045923710836
-1.34309815700804 1.29796349165808
-1.33341170594943 1.32527576474472
-1.3231699173536 1.35238468138903
-1.31237705670707 1.3792789513054
-1.30103761900612 1.40594737360399
-1.28915632688471 1.4323788414557
-1.2767381286476 1.45856234671786
-1.26378819620952 1.48448698451898
-1.25031192294114 1.51014195780031
-1.23631492142285 1.53551658181263
-1.22180302110727 1.56060028856623
-1.20678226589138 1.58538263123222
-1.19125891159937 1.60985328849341
-1.17523942337723 1.63400206884295
-1.15873047300015 1.65781891482887
-1.14173893609387 1.68129390724276
-1.12427188927111 1.70441726925095
-1.10633660718436 1.72717937046631
-1.08794055949609 1.74957073095916
-1.06909140776782 1.7715820252054
-1.04979700226924 1.79320408597042
-1.03006537870873 1.81442790812706
-1.0099047548867 1.83524465240603
-0.989323527272975 1.85564564907732
};
\end{axis}

\end{tikzpicture}
    }
    \caption{}
    \label{fig:ball_hit_scatter}
     \end{subfigure}
     \vspace{5pt}
        \caption{Fig.~\ref{fig:system-reachibility} shows a  histogram of successful returns as a function of the distance the wheelchair needed to move to hit the ball. Fig.~\ref{fig:intercept_time_hist} shows a histogram for times of ball arrival at interception point from the time of the first detection with the court setup. Fig.~\ref{fig:ball_hit_scatter} shows a scatter plot of ball positions in the interception plane with the court setup.}
        \label{fig:combined_new}
\end{figure*}

To further evaluate the effect of variance in ball launches on the robot's performance, we conducted a contiguous trial of 50 launches with a ball launcher in the court setting and plotted the histogram of the time it takes to reach the pre-specified interception plane \SI{8}{\meter} away from the launcher in Fig.~\ref{fig:intercept_time_hist}.
In Fig.~\ref{fig:ball_hit_scatter}, we visualize the position of the balls at the interception plane and the configurations that were reachable without wheelchair movement. 65\% of the launched balls were in a configuration that required wheelchair movement. The experiment was repeated in the lab setting with better results due to the closer proximity of the cameras and more controlled conditions. These results along with the data in Table~\ref{tab:results} serve as a baseline for future work.

\section{Discussion and Key Challenges}
\label{sec:discussion}

We have experimentally demonstrated that ESTHER is fast, agile, and powerful enough to successfully hit ground strokes on a regulation tennis court. Our vision for the system is to be able to rally and play with a human player. To realize that goal, we lay down a set of key challenges to address. 

Section~\ref{subsec:reachability} showcased that ESTHER is able to successfully return balls that require the wheelchair to move up to \SI{2}{\meter} given that the movement starts right after the ball is launched. However, as discussed in Section~\ref{subsec:sensing-state-estimation-capabilities}, our vision system needs some time to converge to an accurate prediction. To get an accurate, early estimate of the ball's pose, we will need a higher fps vision setup that utilizes cameras to triangulate a ball from multiple angles~\cite{jonas-tabletenniskukaspin2020}. To make the ball's predicted trajectory more precise, the system should also be able to estimate the spin of the ball. Directly estimating spin on the ball is a hard problem~\cite{jonas-tabletenniskukaspin2020}. However, the ball's spin can be estimated indirectly using methods such as trajectory fitting~\cite{jonas-tabletenniskukaspin2020} or an AEKF~\cite{AdaptiveEKF_mobilerobot}. An interesting research direction would be to accurately estimate the ball's state~\cite{pose_trajectoryprediction} by observing how the ball was hit on the previous shot for e.g., a racket going from low to high indicates top spin, and a racket going from high to low is indicative of backspin. Additionally, the stroke of the opposing player can be used to infer the general direction of the ball trajectory enabling early movement of wheelchair. The ultimate challenge is to perform this sensing with only onboard mechanisms.

We currently plan the motion of the wheelchair and the arm separately. However, the two components are dynamically coupled, and an ideal planner should consider both simultaneously. In future work, our system serves as an ideal test-bed for the development of  kinodynamic motion planners for athletic mobile manipulators as current dynamical planners are not good enough for agile movement~\cite{ros_giralt_2021}. Such a planner would also help us reach higher racket head velocities as we will be able to utilize the wheelchair's potential to rotate at high angular speeds as established in the same way that professional human players rotate their trunk while hitting a ground stroke. Moreover, the planner will need to be safe and fast for live human-robot play.

\vspace{-1mm}
\section{Future Work}
\label{sec:applications}

The system opens up numerous exciting opportunities for future research such as learning strokes from expert demonstrations via imitation \& reinforcement learning, incorporating game knowledge to return at strategic points, and effective human-robot teaming for playing doubles. We briefly describe our vision for some of these exciting future directions:

\textbf{Interactive Robot Learning} -- Interactive Robot Learning (i.e., Learning from Demonstration (LfD)) focuses on extracting skills from expert demonstrations or instructions for how to perform a task. It is quite challenging to manually encode various strategies, tactics, and low-level stroke styles for a tennis-playing robot, and, ESTHER  serves as a great platform to develop and deploy LfD algorithms that learn from expert gameplay. Prior work has shown how kinesthetic teaching can be employed to teach robots various strokes for playing Table-Tennis~\cite{chen-irltabletennis2020,gomez-promptabletennis2016,chen2021learning}, but these techniques are demonstrated on arms that are mounted to a stationary base and require much lower racket head speeds. Therefore, we challenge the LfD community to leverage our platform to explore a more challenging robotics domain.

\textbf{Human-Robot Teaming} -- Agile robots that collaborate well with humans on a shared task is a challenging problem as it requires an accurate perception of human intention, clear communication of robot intention, and collaborative planning and execution, where tennis doubles is one such setting~\cite{lee_arjun_2023effect_robotskill}. The small mobile form factor of ESTHER coupled with our portable decentralized vision system, offers an attractive platform that we aim to extend for doubles play. While our system currently lacks the sensing and perception mechanisms to ensure human safety for doubles play, we aim to improve these aspects of our system to enable the study of fast-paced human-robot collaboration. Building trust and personalized strategies between robots and humans over time in a highly competitive and dynamic environment is an important research problem to explore~\cite{hri_survey1}.

\textbf{Robot Hardware Design} --  Finally, we note that there is an inherent trade-off between leveraging existing, commerical-off-the-shelf parts and designing custom components for a novel robot system. This is highlighted by the fact that even with the robot operating at full speed, the ball does not land past the baseline on the opponent's side. Therefore, we propose to explore new, custom hardware designs to improve the performance of ours system as inspired by prior work in designing muscular robots for table tennis~\cite{buchler2022learning}.

\section{Conclusion} 
\label{sec:conclusion}

In this paper, we present ESTHER for robotic tennis. Our work contribution is threefold: introduction of a low-cost, fast, decentralized perception system that can be easily set up anywhere for accurate ball tracking; design of a chain drive system that can be used to motorize a regulation sports wheelchair into a highly agile mobile base that is fit for use in highly dynamic settings; and planning and control of an agile mobile manipulator to exhibit athletic behaviors. We experimentally evaluated the capabilities of individual subsystems and demonstrated our system can successfully hit ground strokes on a regulation tennis court. Future research includes improvements to the perception system, kinodynamic planners, learning game strategies, court positioning, and strokes from human play, as well as safety improvements to enable human-robot collaboration in doubles tennis.

\vspace{-2mm}
\bibliographystyle{IEEEtran}
\bibliography{references}

\end{document}